\documentclass[11pt,a4paper]{article}
\usepackage[hyperref]{emnlp2021}

\usepackage{times}
\usepackage{latexsym}
\usepackage{url}  
\usepackage{graphicx}  
\usepackage{caption}
\usepackage{xcolor} 
\usepackage{colortbl} 
\usepackage{booktabs}
\usepackage{dirtytalk}
\usepackage{amsmath}
\usepackage{amssymb}
\usepackage{hyperref}
\usepackage{wrapfig}
\usepackage{array,multirow}
\usepackage{algorithm}
\usepackage{algpseudocode}
\usepackage{nicefrac}       
\usepackage{microtype}      
\usepackage{subcaption}
\usepackage{xfrac}
\usepackage{mathtools}
\usepackage{soul}
\usepackage{tabularx}

\usepackage{multirow}
\usepackage{color}
\usepackage{mathtools}

\usepackage{defs}

\usepackage[english]{babel}
\usepackage{hyperref}
\addto\captionsenglish{%
}
\addto\extrasenglish{%
}

\usepackage{xspace}
\usepackage{CJKutf8}
\usepackage{xtab,booktabs}

\newcommand{\dataset}[1]{\textsl{#1}}
\newcommand{\english}{\dataset{EN-ext}\xspace}
\newcommand{\englishclo}{\dataset{EN}\xspace}
\newcommand{\german}{\dataset{DE}\xspace}
\newcommand{\germant}{\dataset{DE-EN}\xspace}
\newcommand{\japanese}{\dataset{JP}\xspace}
\newcommand{\japaneset}{\dataset{JP-EN}\xspace}
\newcommand{\semeval}{\dataset{SemEval}\xspace}

\newcommand{\combw}{\dataset{Comb}\xspace}

\newcommand{\mbert}{\textbf{mBERT}\xspace}
\newcommand{\xlmrob}{\textbf{XLM-RoBERTa}\xspace}
\newcommand{\xlm}{\textbf{XLM-w/o-Emb}\xspace}

\newcommand{\best}[1]{\textbf{#1}\xspace}
\newcommand{\secbest}[1]{\textit{\textbf{#1}}\xspace}

\title{\emph{I Wish I Would Have Loved This One, But I Didn't} -- A Multilingual Dataset for Counterfactual Detection in Product Reviews}

\author{James O'Neill$^{\ddagger}$\thanks{The two first authors contributed equally} \\ James.O-Neill@liverpool.ac.uk \And Polina Rozenshtein$^{\dagger,*}$ \\ prrozens@amazon.co.jp \And Ryuichi Kiryo$^{\dagger}$ \\ kiryor@amazon.co.jp \AND  Motoko Kubota$^{\dagger}$ 
\\ kubmotok@amazon.co.jp
\\ {\kern 14em} $\text{Amazon}^{\dagger}, \text{University of Liverpool}^{\ddagger}$ 
\And Danushka Bollegala$^{\dagger, \ddagger}$
\\ danubol@amazon.com}

\date{}

\begin{document}
\maketitle
\begin{abstract}
Counterfactual statements describe events that did not or cannot take place.
We consider the problem of counterfactual detection (CFD) in product reviews.
For this purpose, we annotate a multilingual CFD dataset from Amazon product reviews covering counterfactual statements written in English, German, and Japanese languages.
The dataset is unique as it contains counterfactuals in multiple languages, covers a new application area of e-commerce reviews, and provides high quality professional annotations. 
We train CFD models using different text representation methods and classifiers.
We find that these models are robust against the selectional biases introduced due to cue phrase-based sentence selection.
Moreover, our CFD dataset is compatible with prior datasets and can be merged to learn accurate CFD models.
Applying machine translation on English counterfactual examples to create multilingual data performs poorly, demonstrating the language-specificity of this problem, which has been ignored so far.
\end{abstract}

\section{Introduction}

Counterfactual statements are an essential tool of human thinking and are often found in natural languages.
Counterfactual statements may be identified as statements of the form -- \emph{If $p$ was true, then $q$ would be true} (i.e. assertions whose antecedent ($p$) and consequent ($q$) are known or assumed to be false)~\cite{milmed1957counterfactual}. 
In other words, a counterfactual statement describes an event that may not, did not, or cannot take place, and the subsequent consequence(s) or alternative(s) did not take place.
For example, consider the counterfactual statement -- \emph{\textcolor{blue}{I would have been content with purchasing this iPhone}, \textcolor{red}{if it came with a warranty!}}. 
Counterfactual statements can be broken into two parts: a statement about the event (\emph{\textcolor{red}{if it came with a warranty}}), also referred to as the \textbf{antecedent}, and the consequence of the event (\emph{\textcolor{blue}{I would have been content with purchasing this iPhone}}), referred to as the \textbf{consequent}.
Counterfactual statements are ubiquitous in natural language and have been well-studied in fields such as philosophy~\cite{lewis2013counterfactuals}, psychology~\cite{markman2007implications, roese1997counterfactual}, linguistics~\cite{ippolito2013counterfactuals}, logic~\cite{milmed1957counterfactual,quine1982methods}, and causal inference~\cite{hofler2005causal}. 

Accurate detection of counterfactual statements is beneficial to numerous applications in natural language processing (NLP) such as in medicine (e.g., clinical letters), law (e.g., court proceedings), sentiment analysis, and information retrieval.
For example, in information retrieval, counterfactual detection (CFD) can potentially help to remove irrelevant results to a given query.
Revisiting our previous example, we should not return the iPhone in question for a user who is searching for \emph{iPhone with warranty} because that iPhone does not come with a warranty. 
A simple bag-of-words retrieval model that does not detect counterfactuals would return the iPhone in question because all the tokens in the query (i.e. \emph{iPhone}, \emph{with}, \emph{warranty}) occur in the review sentence.
Detecting counterfactuals can also be a precursor to capturing causal inferences~\cite{wood2018challenges} and interactions, which have shown to be effective in fields such as health sciences~\cite{hofler2005causal}. 
\newcite{janocko2016counterfactuals} and \newcite{son2017recognizing} studied CFD in social media for automatic psychological assessment of large populations. 

CFD is often modelled as a binary classification task~\cite{son2017recognizing,yang2020semeval}.
A manually annotated sentence-level counterfactual dataset was introduced in SemEval-2020~\cite{yang2020semeval} to facilitate further research into this important problem.
However, successful developments of classification methods require extensive high quality labelled datasets.
To the best of our knowledge, currently there are only two labelled datasets for counterfactuals: (a) the pioneering small dataset of tweets~\cite{son2017recognizing} and (b) a recent larger corpus covering the area of the finance, politics, and healthcare domains~\cite{yang2020semeval}. 
However, these datasets are limited to the English language.

In this paper, we contribute to this emerging line of work by annotating a novel CFD dataset for a new domain (i.e. product reviews), covering languages in addition to English, such as Japanese and German, ensuring a balanced representation of counterfactuals and the high quality of the labelling.
Following prior work, we model counterfactual statement detection as a binary classification problem, where given a sentence extracted from a product review, we predict whether it expresses a counterfactual or a non-counterfactual statement. 
Specifically, we annotate sentences selected from Amazon product reviews, where the annotators provided sentence-level annotations as to whether a sentence is counterfactual with respect to the product being discussed. 
We then represent sentences using different encoders and train CFD models using different classification algorithms.

The percentage of sentences that contain a counterfactual statement in a random sample of sentences has been reported to be low as 1-2\%~\cite{son2017recognizing}.
Therefore, all prior works annotating CFD datasets have used clue phrases such as \emph{I wished} to select candidate sentences that are likely to be true counterfactuals, which are then subsequently annotated by human annotators~\cite{yang2020semeval}.
However, this selection process can potentially introduce a selection bias towards the clue phrases used.

To the best of our knowledge, while the data selection bias is a recognised problem in other NLP tasks~(e.g.,~\newcite{larson2020iterative}), this selection bias on CFD classifiers has not been studied previously.
Therefore, we train counterfactual classifiers with and without masking the clue phrases used for candidate sentence selection. 
Furthermore, we experiment with enriching the dataset with sentences that do not contain clue phrases but are semantically similar to the ones that contain clue phrases.
Interestingly, our experimental results reveal that compared to the lexicalised CFD such as bag-of-words representations, CFD models trained using contextualised masked language models such as BERT are robust against the selection bias~\cite{devlin-etal-2019-bert}. 
Our contributions in this paper are as follows:
\paragraph{First-ever Multilingual Counterfactual Dataset:} 
We introduce the first-ever multilingual CFD dataset containing manually labelled product review sentences covering English, German, and Japanese languages.\footnote{\url{https://github.com/amazon-research/amazon-multilingual-counterfactual-dataset}} 
As already mentioned above, counterfactual statements are naturally infrequent. 
We ensure that the positive (i.e. counterfactual) class is represented by at least 10\% of samples for each language.
Distinguishing between a counterfactual and non-counterfactual statements is a fairly complex task even for humans. 
Unlike previous works, which relied on crowdsourcing, we employ professional linguists to produce a high quality annotation.
We follow the definition of counterfactuals used by \newcite{yang2020semeval}  to ensure that our dataset is compatible with the SemEval-2020 CFD dataset (\textbf{SemEval}). 
We experimentally verify that by merging our dataset with the SemEval CFD dataset, we can further improve the accuracies of counterfactual classifiers. 
Moreover, applying machine translation on the English CFD dataset to produce multilingual CFD datasets results in poor CFD models, indicating the language-specificity of the problem that require careful manual annotations.

\paragraph{Accurate CFD Models:} Using the annotated dataset we train multiple classifiers using (a) lexicalised word-order insensitive bag-of-words representations as well as (b) contextualised sentence embeddings. 
We find that there is a clear advantage to using contextualised embeddings over non-contextualized embeddings, indicating that counterfactuals are indeed context-sensitive.

\section{Related Work}
\label{sec:related}

Counterfactuals have been studied in various contexts such as for problem solving~\cite{markman2007implications}, explainable machine learning~\cite{byrne2019counterfactuals}, advertisement placement~\cite{joachims2016counterfactual} and algorithmic fairness~\cite{kusner2017counterfactual}.
\newcite{kaushik2020learning} proposed an annotation scheme whereby the original data is augmented in a counterfactual manner to overcome spurious associations that a classifier heavily relies upon, thus failing to perform well on test data distributions that are not identical. Unlike~\citet{kaushik2020learning} and closely related work by ~\citet{gardner2020evaluating}, we are interested in identifying existing counterfacts and filtering these statements to improve search performance.

A CFD task was presented in SemEval-2020 Challenge~\cite{Yang:2020}.
The provided dataset contains counterfactual statements from news articles.
However, the dataset does not cover counterfactuals in e-commerce product reviews, which is our focus in this paper.
One of the earliest CFD datasets was annotated by~\newcite{son2017recognizing} and covers counterfactual statements extracted from social media. 
Both datasets are labelled for binary classification by crowdsourcing and contain only sentences in English.  
We will compare our dataset to these previous works in \autoref{datacomparison}. 
To summarise, our dataset is unique as it contains counterfactuals in multiple languages, covers a new application area of e-commerce reviews, and provides high quality annotations.

A range of CFD methods was recently proposed in response to the SemEval-2020 challenge~\cite{Yang:2020}. 
Most of the high performing methods~\cite{ding2020hit, fajcik2020but, lu2020iscas, ojha2020iitk, yabloko2020ethan} use state-of-the-art pretrained language models~\cite{devlin-etal-2019-bert, liu2019roberta, lan2019albert, radford2019language, yang2019xlnet}. 
Traditional ML methods, such as SVM and random forests were also used but with less success~\cite{ojha2020iitk}.

To achieve the best prediction quality, ensemble strategies are employed. 
The top performing systems use an ensemble of transformers~\cite{ding2020hit, fajcik2020but, lu2020iscas}, while others include Convolutional Neural Networks (CNNs) with Global Vectors~\cite[GloVe;][]{Pennington:EMNLP:2014} embeddings~\cite{ojha2020iitk}.
Various structures are used on top of transformers. For example, \newcite{lu2020iscas, ojha2020iitk} use a CNN as the top layer, while \newcite{bai2020byteam} use a Bi-GRUs and Bi-LSTMs. 
Some other proposed methods use additional modules, such as constituency and dependency parsers, in the lower layers of the architecture~\cite{yabloko2020ethan}.

CFD datasets tend be highly imbalanced because counterfactual statements are less frequent in natural language texts.
Prior work has used techniques such as pseudo-labelling~\cite{ding2020hit} and multi sample dropout~\cite{chen2020ferryman} to address the data imbalance and overfitting problems.


\section{Dataset Curation}
\label{sec:curation}

We adopt the definition of a counterfactual statement proposed by \newcite{janocko2016counterfactuals} where they define it as \emph{a statement which looks at how a hypothetical change in past experience could have affected the outcome of that experience.}
Their definition is based on linguistic structures of 6 types of counterfactuals as following.

{\bf Conjunctive Normal:} The antecedent is followed by the consequent.
The antecedent consists of a conditional conjunction followed by a past tense subjunctive
verb or past modal verb.
The consequent contains a past or present tense modal verb.
({\it Example: \textbf{If} everyone \textbf{got} along, it would be more enjoyable.}) 

{\bf Conjunctive Converse:} The consequent is followed by the antecedent.
The consequent consists of a modal verb and past or present tense verb.
The antecedent consists of a conditional conjunction followed by a past tense subjunctive
verb or past tense modal.
({\it Example: I would be stronger, \textbf{if} I \textbf{had lifted} weights.})

{\bf Modal Normal:} The antecedent is followed by the consequent.
The antecedent consists of a modal verb and past participle verb.
The consequent consists of a past/present tense modal verb.
({\it Example: We \textbf{should have gone} bowling, that \textbf{would have been} better.})

{\bf Wish/Should Implied:} The antecedent is present, the consequent is implied.
The antecedent is the independent clause following `wish' or `should'.
The consequent is implied and can be paraphrased as ``would be better off''.
({\it Examples: I \textbf{wish I had} been richer. I \textbf{should have} revised my rehearsal lines.})

{\bf Verb Inversion:} No specific order of the antecedent and consequent.
The antecedent uses the subjunctive mood by inverting the verbs `had' and `were' to create a
hypothetical conditional statement along with a past tense verb.
The consequent consists of a modal verb and past or present tense verb.
({\it Example: \textbf{Had I listened} to your advice, I \textbf{may have got} the job.})

{\bf Modal Propositional, Would/Could Have:} The consequent is followed by the antecedent.
The antecedent consists of a past/present modal verb.
The consequent consists of a prepositional phrase (only certain types).
({\it Examples:	I \textbf{would have been better off} not reading this. I \textbf{would have been happier} without John.})

Note that, while \newcite{yang2020semeval} explicitly mention only $5$ types of counterfactual and~\newcite{son2017recognizing} work with $7$ types, their definitions and clue words used for data collection effectively cover the same $6$ types defined by~\newcite{janocko2016counterfactuals}. 
We worked with professional linguists to extend these counterfactual definitions for the German and Japanese languages. 
While the extension of the definition from English to German is relatively straightforward, the extension to syntactically and orthographically different structure of Japanese sentences was challenging~\cite{jacobsen} and required re-writing the annotation guidelines including additional examples. 
The annotation guidelines are included in the dataset release. 

\subsection{Data Collection}
\label{sec:collection}

The main step of data collection in the previous works~\cite{son2017recognizing, yang2020semeval} is filtering of the data using a pre-compiled list of clue words/phrases. 
Because the exact list of clue phrases used by~\newcite{janocko2016counterfactuals} was not publicly available, we created a new list of clue phrases following the definitions of counterfactual types. 
In addition, we compiled similar clue phrase lists for German and Japanese languages.
\newcite{yang2020semeval} applied a more complex procedure, where they match Part of Speech (PoS)-tagged sentences against lexico-syntactic patterns. 
In our work, we do not consider PoS-based patterns, which are difficult to generalise across languages.

We use the Amazon Customer Reviews Dataset,\footnote{\url{https://s3.amazonaws.com/amazon-reviews-pds/readme.html}} which contains over 130 million customer reviews collected and released by Amazon to the research community.
To create an annotated dataset, we select reviews in different categories as detailed in the Supplementary.
Next, we sample candidate sentences for annotation in two iterations.

In the first iteration, we consider reviews written by customers with a verified purchase (i.e., the customer has bought the product about which he or she is writing the review).
Given that counterfactual statements are infrequent,
all prior works~\cite{son2017recognizing, yang2020semeval}
have used clue phrase lists for selecting data for human annotation.
Following this practice, we select sentences that contain exactly one clue phrase from our pre-compiled clue phrase lists for each language.
We remove sentences that are exceedingly long (more than 512 tokens) or short (less than 10 tokens). 
Shorter sentences might not contain sufficient information for a human annotator to decide whether it is a counterfactual statement, whereas longer sentences are likely to contain various other information besides counterfactuals.

The above-mentioned first iteration might produce a biased dataset in the sense that all sentences contain counterfactual clues from the predefined lists.
There are two possible drawbacks in this selection method.
First, the manually compiled clue phrase lists might not cover all the different ways in which we can express a counterfactual in a particular language.
Therefore, the sentences selected using the clue phrase lists might have coverage issues.
Second, a counterfactual classification model might assign high confidence scores for some high precision clue phrases (e.g., ``wish'' for English).
Such a classifier is likely to perform poorly on test data that do not use clue phrases for expressing counterfactuality.
On the contrary, adding sentences with no clue words to the dataset might result in a greater bias: those additional sentences are likely to be negative examples, and thus discriminatory power of the clue phrases can get amplified.
Later in our experiments, we empirically evaluate the effect of selection bias due to the reliance on clue phrases.

To address the selection bias, in addition to the sentences selected in the first iteration, we conduct a second iteration where we select sentences that \emph{do not} contain counterfactual clues from our lists.
For this purpose, we create sentence embeddings for each sentence selected in the first iteration. We use a pretrained multilingual BERT model\footnote{\url{https://huggingface.co/bert-base-multilingual-uncased}}.
We then use $k$-means clustering to cluster these sentences into $k=100$ clusters.
We assume each cluster represents some aspect of a product, and represented by its centroid. 
Next, we pick sentences that do not contain the clue phrases, compute their sentence embeddings, and measure the similarity to each of the centroids. For each centroid we select the top $n$ most similar sentences for manual annotation.
We set $n$ such that we obtain an approximately equal number of sentences to the number of sentences that contain clue phrases selected in the first iteration.
All selected sentences are manually annotated for counterfactuality as described in \autoref{sec:annotate}.

\subsection{Annotation}
\label{sec:annotate}

The annotators were provided guidelines with definitions, extensive examples and counterexamples.
Briefly, counterfactual statements were identified if they belong to any of the counterfactual types described in \autoref{sec:curation}.
If any part of a sentence contains a counterfactual, then we consider the entire sentence to be a counterfactual. 
This annotation process increases the number of counterfactual examples and the coverage across the counterfactual types in the dataset, thereby improving the class imbalance.
 We require that at least $90\%$ of the sentences have agreement of 2 professional linguists (2 out of 2 agreement), the rest at most $10\%$ cases had a third linguist to resolve the disagreement (2 out of 3 agreement). 

\subsection{Dataset Statistics}
The basic dataset statistics can be found in \autoref{tab:stats}. 
We present two versions of the English dataset: \englishclo contains only sentences filtered by the clue words, \english is a superset of \englishclo enriched by sentences with no clue words as described above. The clue-based dataset \englishclo contains about 1/5-th of positive examples, while its extended version contains 1/10-th of counterfactuals. Only 76 out of 4977 added sentences were labelled positively.
\german dataset contains $69.1\%$ and \japanese contains $9.5\%$ of counterfactuals.

The summary of clue phrase distributions in positive and negative classes is shown in \autoref{tab:sum:clue_words}. 
Interestingly, English and German lists have approximately the same number of clues, but the precision for German clues is much higher, resulting in more counterfactual statements being extracted using those clue phrases.
On the contrary, the Japanese list has the largest number of clues, yet results in the lowest precision. 
The specification of counterfactual clue phrases for Japanese is a linguistically hard problem because the meaning of the clues is highly context dependent. 
The large number of Japanese clue phrases is due to the orthographic variations present in Japanese where the same phrase can be written using kanji, hiragana, katakana characters or a mixture of them.
Because we were able to select a sufficiently large datasets for German and Japanese using the clue phrases, we did not consider the second iteration step described in \autoref{sec:collection} for those languages.

\begin{table}[t]
\small
	\setlength{\tabcolsep}{0pt}
	\centering
	\begin{tabular*}{0.45\textwidth}{@{\extracolsep{\fill}} lcccc}
		\toprule 
		 Dataset &Positive & Negative & Total & CF \%\\
		 \midrule
		 \englishclo & 954 & 4069 & 5023 & 18.9\\
		 \english & 1030 & 8970 &10000 & 10.0\\
		 \german & 4840 & 2160 & 7000 & 69.1\\
		 \japanese & 667 & 6333 & 7000& 9.5\\
\bottomrule
\end{tabular*}
\caption{Dataset statistics: the number of positive (counterfactual) and negative (non-counterfactual) examples, total sizes of the datasets, percentage of counterfactual (CF) examples.}
\label{tab:stats}
\end{table}

\begin{table}[t]
\small
	\setlength{\tabcolsep}{0pt}
	\centering
	\begin{tabular*}{0.45\textwidth}{@{\extracolsep{\fill}} lcccc}
		\toprule 
		Dataset& $N$ & $f_{P}$&$f_{N}$ & $f_{data}$  \\
		\midrule
		\englishclo & 29 & 100. & 100. & 100.\\
		\english & 29 &92.6&45.3&50.2\\
		\german & 27 &100.&100.&100.\\
		\japanese &70 &100.&100.&100.\\
		\bottomrule
	\end{tabular*}
	\caption{Clue phrases summary for the datasets: $N$ is the total number of clue phrases in each clue phrase list. $f_P$ and $f_N$ are the percentages of examples containing clue phrases respectively in counterfactual and non-counterfactal classes. $f_{data}$ is the percentage of sentences containing a clue phrase in a dataset.}
	\label{tab:sum:clue_words}
\end{table}

\subsection{Comparison with Existing Datasets}
\label{datacomparison}

We compare the multilingual counterfactual dataset we create against existing datasets in \autoref{tab:comp}.
Our dataset is well-aligned with the two other existing datasets in the sense that we use the same definition of a counterfactual, keep a similar percentage of positive examples, and use similar keywords for dataset construction. 
These properties ensure that our dataset of product reviews can be used on its own, as well as organically combined with the existing datasets from other domains. 
A distinctive feature of our dataset is its coverage of a novel domain, e-commerce reviews, which is not covered by any of the existing counterfactual datasets. 
Furthermore, our dataset is available for three languages: English, German, and Japanese. 
This is the first counterfactual dataset not limited to English language. 
Unlike previous works, which relied on crowdsourcing, we employ professional linguists to produce the lists of clue words and supervise the annotation. This ensures the high quality of the labelling.

\begin{table*}[t]
    \small
	\setlength{\tabcolsep}{0pt}
	\centering
	\begin{tabular*}{\textwidth}{@{\extracolsep{\fill}} llll}
		\toprule 
Dataset &  Language & Size & CF \%\\
\midrule
\newcite{son2017recognizing} &  English & 1637 (2137) & 10.1 (31.2)\\ 
\newcite{yang2020semeval} & English & 20000 & 11.0\\
This work  & English / German / Japanese & 10000 (5023) / 7000 / 7000 & 10.0 (18.9) / 69.1 / 9.5 \\
\end{tabular*}
\begin{tabular*}{\textwidth}{@{\extracolsep{\fill}} p{2.4cm}p{2cm}p{3cm}p{3cm}p{4.5cm}}
		\toprule 
		Dataset & CF definition & Domain & Construction & Annotation \\
		\midrule
		\newcite{son2017recognizing} & \newcite{janocko2016counterfactuals} & Twitter & keywords filtering &
		mixed: manual (unknown), automatic pattern matching\\ 
		\newcite{yang2020semeval} & \newcite{janocko2016counterfactuals} & News: finance, politics, healthcare &keywords filtering, pattern matching & manual (crowdsourcing, strong agreement) \\	
		This work & \newcite{janocko2016counterfactuals}& Amazon Reviews & keywords filtering  & manual (curated by linguists)  \\
		\bottomrule
	\end{tabular*}
	\caption{Dataset comparisons. The numbers in parenthesis for \newcite{son2017recognizing} correspond to the union of manually and automatically labelled datasets. The numbers in parenthesis for this work correspond to clue-based English dataset \englishclo.}
		\label{tab:comp}
\end{table*}

\section{Evaluations}
\label{sec:classifiers}

We conduct a series of experiments to systematically evaluate several important factors related to counterfactuality such as (a) selection bias due to clue phrases (\autoref{sec:clues}), 
(b) effect of merging multiple counterfactual datasets (\autoref{sec:merge}),
(c) use of machine translation (MT) to translate counterfactual statements (\autoref{sec:trans}), and
(d) effect of different sentence encoders and classifiers for training CFD models (\autoref{sec:class}).

For evaluations in (a), (b), and (c), we fine-tune a widely used multilingual transformer model BERT (mBERT)~\cite{devlin-etal-2019-bert} to train a CFD model.
The model is pretrained for the tasks of masked language modelling and next sentence prediction for 104 languages\footnote{\url{https://huggingface.co/bert-base-multilingual-uncased}} and is used with the default parameter settings.
The model is implemented using the Transformer.\footnote{\url{https://github.com/huggingface/transformers}} library
We fine-tune a linear layer on top of these pretrained language models for the CFD task using the training process as described next.\footnote{See Supplementary for the details on fine-tuning.}

We use an 80\%-20\% train-test data split and tune hyperparameters via   5-fold cross-validation.
Hyperparameters in the already pretrained transformer models are kept fixed.
F1, Matthew's Correlation Coefficient~\citep[MCC;][]{boughorbel2017optimal}, and accuracy are used as evaluation metrics. 
MCC ($\in [-1, 1]$) accounts for class imbalance and incorporates all correlations within the confusion matrix~\cite{chicco2020advantages}. 
Accuracy may be misleading in highly imbalanced datasets because a
simple classification of all instances to the majority class has a high accuracy.
However, for consistency with prior work, we report all three evaluation metrics in this paper.
All the reported results are averaged over at least $3$ independently trained models initialised with the same hyperparameter values. 
For tokenisation, unless the tokeniser is pre-specified for the model, we use {\small\texttt{word\_tokenize}} from {\small\texttt{nltk.tokenize.punkt}}\footnote{\url{https://www.nltk.org/api/nltk.tokenize.html}} for English and German languages; and MeCab\footnote{\url{https://pypi.org/project/mecab-python3/}} as the morphological analyser for Japanese.

\subsection{Selection Bias due to Clue Phrases}
\label{sec:clues}

To evaluate the effectiveness of clue phrases for selecting sentences for human annotation and any selection bias due to this process, we fine-tune mBERT with and without masking the clue phrases.
Classification performance values are shown in \autoref{tab:masked}.
Overall, we see that \textbf{no mask} (training without masking) returns slightly better performance than \textbf{mask}  (training with masking), however the differences are not statistically significant.
This is reassuring because it shows that the sentence embeddings produced by mBERT generalise well beyond the clue phrases used to select sentences for manual annotation.
On the other hand, if a CFD model had simply \emph{memorised} the clue phrases and was classifying based on the occurrences of the clue phrases in a sentence, we would expect a drop in classification performance in \textbf{no mask} setting due to overfitting to the clue phrases that are not observed in the test data.
Indeed for \englishclo where all sentences contain clue phrases, we see a slight drop in all evaluation measure for \textbf{no mask} relative to \textbf{mask}, which we believe is due to this overfitting effect.
The performance on \japanese is the lowest among all languages compared.
This could be attributed to the tokenisation issues and lack of Japanese coverage in mBERT. 
Many counterfactual clues in Japanese are parts of verb/adjective inflections, which can get split/removed during the tokenisation. 

\begin{table}[t]
    \small
	\setlength{\tabcolsep}{0pt}
	\centering
	\begin{tabular*}{0.48\textwidth}{@{\extracolsep{\fill}} llccc}
			\toprule
			Dataset & Mask &  \multicolumn{3}{c}{\textbf{mBERT}} \\
			&& F1 & MCC & Acc \\
			\midrule
			\multirow{2}{*}{\englishclo}&mask & 0.92 & 0.76& 0.92\\
			&no mask & 0.89 & 0.73 & 0.89\\
			\midrule
			\multirow{2}{*}{\english}&mask & 0.93 & 0.69& 0.93\\
			&no mask & 0.94 & 0.74 & 0.94\\
			\midrule
			\multirow{2}{*}{\german}&mask & 0.86 & 0.68& 0.86\\
			&no mask & 0.90 & 0.79 & 0.90\\
			\midrule
			\multirow{2}{*}{\japanese} & mask & 0.86 & 0.48 & 0.84\\
			& no mask & 0.85 & 0.49 & 0.82\\
			\bottomrule
	\end{tabular*}%
	\caption{F1, MCC and Accuracy (Acc) for CFD models trained with and without masking the clue phrases.}
	\label{tab:masked}
\end{table}

\autoref{tab:metrics} shows recall ($R$) and precision ($P$) on masked (subscript ${m}$) and non-masked (subscript ${nm}$) settings.
In all datasets the recall is higher than precision for both masked and non-masked versions due to dataset imbalance with an underrepresented positive class. 
The number of positive examples misclassified under masked and non-masked settings are typically very small. 
We see that the CFD model trained on \english has a higher recall, but lower precision than the one on \englishclo. Most of the added examples in \english are negatives, which makes it hard to maintain a high precision.

\begin{table}[t]
\small
	\setlength{\tabcolsep}{0pt}
	\centering
	\begin{tabular*}{0.48\textwidth}{@{\extracolsep{\fill}} lllll}
		\toprule
Metric & \englishclo & \english & \german & \japanese \\
\midrule
$R_{nm}$ & 0.93 & 0.94 & 0.92 & 0.85\\
$P_{nm}$ & 0.71 & 0.59 & 0.94 & 0.30\\
$R_{m}$ & 0.87 & 0.79 & 0.86 & 0.88\\
$P_{m}$ & 0.68 & 0.66 & 0.93 & 0.37\\
\bottomrule
	\end{tabular*}%
	\caption{Precision and Recall for mBERT trained with ($m$) and without  ($nm$) masking the clue phrases.}
	\label{tab:metrics}
\end{table}

\subsection{Cross-Dataset Adaptation}
\label{sec:merge}

To study the compatibility of our dataset with existing datasets, we train a CFD model on one dataset and test the trained model on a different dataset.
Prior work on domain adaptation~\cite{Blitzer:ML:2009} has shown that the classification accuracy of such a cross-domain classifier is upper-bounded by the similarity between the train and test datasets.
Further, we merge our \english dataset with the \semeval dataset~\cite{yang2020semeval} to create a dataset denoted by \combw.
Specifically, we separately pool the the counterfactual and noncounterfactual instances in each dataset to create \combw.

As can be seen from \autoref{tab:combined}, the models trained on \englishclo and \english perform poorly on \semeval, while the model trained on \semeval has relatively high values of F1, MCC, and Accuracy on \englishclo and \english. 
This implies that the product reviews we use cover a narrow subdomain compared to the domains in \semeval. 
Interestingly, the CFD model trained on \combw reports the best performance across all measures, indicating that our dataset is compatible with \semeval and can be used in conjunction with existing datasets to train better CFD models.

\begin{table}[t]
\small
\setlength{\tabcolsep}{0pt}
\centering
\begin{tabular*}{0.48\textwidth}{@{\extracolsep{\fill}} llccc}
\toprule
Train & Test &  \multicolumn{2}{c}{\textbf{mBERT}} \\
& & F1 & MCC & Acc \\
\midrule
\multirow{4}{*}{\englishclo}
&\englishclo&0.89 & 0.73 & 0.89\\
&\english&0.96 & 0.85 & 0.96\\
&\semeval&0.65 & 0.28 & 0.59\\
&\combw&0.68 & 0.31 & 0.62\\
\midrule
\multirow{4}{*}{\english}
&\englishclo&0.92 & 0.80 & 0.92\\
&\english&0.94 & 0.74 & 0.94\\
&\semeval&0.50 & 0.19 & 0.42\\
&\combw&0.49 & 0.19 & 0.42\\
\midrule
\multirow{4}{*}{\semeval}
&\englishclo&0.82 & 0.56 & 0.80\\
&\english&0.86 & 0.48 & 0.83\\
&\semeval&0.93 & 0.71 & 0.92\\
&\combw&0.96 & 0.84 & 0.96\\
\midrule
\multirow{4}{*}{\combw}
&\englishclo&0.95 & 0.86 & 0.95\\
&\english&0.94 & 0.72 & 0.94\\
&\semeval&0.93 & 0.70 & 0.92\\
&\combw&0.96 & 0.84 & 0.96\\
\bottomrule
\end{tabular*}%
\caption{Classification quality, combining datasets for training and evaluation.}
\label{tab:combined}
\end{table}

\subsection{Cross-Lingual Transfer via MT}
\label{sec:trans}

Considering the costs involved in manually annotating counterfactual statements for each language, a frugal alternative would be to train a model for English and then apply it on test sentences in a target language of interest, which are translated into English using a machine translation (MT) system.
To evaluate this possibility, we first translate the German and Japanese CFD datasets into English (denoted respectively by \germant and \japaneset) using Amazon MT.\footnote{\url{https://aws.amazon.com/translate/}}
Next, we train separate English CFD models using \englishclo, \english and \semeval datasets, and apply those models on \germant and \japaneset.

As shown in \autoref{tab:translated}, the MCC values for the MT-based CFD model are significantly lower than that for the corresponding in-language baseline, which is trained using the target language data.
Therefore, simply applying MT on test data is \emph{not} an alternative to annotating counterfactual datasets from scratch for a novel target language.
This result shows the importance of developing counterfactual datasets for languages other than English, which has not been done prior to this work.
Moreover, the performance for German, which belongs to the same Germanic language group as English, is better than for Japanese.
The model trained on \semeval performs the worst on \germant dataset, and has the lowest MCC on \japaneset. 
This experimental result indicates the importance of introducing new languages to the counterfactual dataset family.

\begin{table}[t]
\small
	\setlength{\tabcolsep}{0pt}
	\centering
	\begin{tabular*}{0.48\textwidth}{@{\extracolsep{\fill}} lcccc}
		\toprule
		Train & Test &  \multicolumn{3}{c}{\textbf{mBERT}} \\
		&& F1 & MCC& Acc \\
		\midrule
		\englishclo&\germant&0.65&0.41&0.64\\
		\english&\germant&0.73&0.49&0.72\\
		\semeval&\germant&0.58&0.35&0.58\\
		\german&\german&0.90&0.79&0.90\\
		\midrule
		\englishclo&\japaneset&0.80&0.26&0.78\\
		\english&\japaneset&0.80&0.28&0.76\\
		\semeval&\japaneset&0.86&0.22&0.86\\
		\japanese & \japanese & 0.85 & 0.49 & 0.82\\
		\bottomrule
	\end{tabular*}%
	\caption{Classification quality of English translations.}
	\label{tab:translated}
\end{table}

\subsection{Sentence Encoders and Classifiers}
\label{sec:class}

We evaluate the effect of the sentence encoding and binary classification methods on the performance of CFD using multiple settings.
\paragraph{Bag-of-N-grams (BoN):}
We represent a sentence using tf-idf weighted unigrams and bi-grams and ignore $n$-grams with a frequency less than 2 or more than 95\% of the frequency distribution. Next, Principal Component Analysis~\citep[PCA;][]{wold1987principal} is used to create 600-dimensional sentence embeddings.
\paragraph{Word Embeddings (WE):}
We average the 300-dimensional fastText embeddings trained on Common Crawl and Wikipedia\footnote{ \url{https://fasttext.cc/docs/en/crawl-vectors.html}}
for the words in a sentence to create its sentence embedding.
We note that there have been meta-embedding methods~\cite{Bollegala:2018aa,Bollegala:IJCAI:2018} proposed to combine multiple word embeddings to further improve their accuracy.
However, their consideration for CFD is beyond the scope of current work.

BoN and WE representations are used to train binary CFD models using different classification methods such as a Support Vector Machine~\cite[SVM;][]{cortes1995support}
with a Radial Basis function, an ID3 Decision Tree \cite[DT;][]{breiman1984classification}, a Random Forest ~\cite[RF;][]{breiman2001random} with 20 trees.

\paragraph{Pretrained Language Models}
Along with mBERT, we fine-tune a linear layer for CFD task on top of two following pretrained transformer models:
XLM model~\cite{DBLP:conf/nips/ConneauL19}\footnote{\url{https://huggingface.co/xlm-mlm-100-1280}} and 
base XLM-RoBERTa model~\cite{DBLP:conf/acl/ConneauKGCWGGOZ20}.\footnote{\url{https://huggingface.co/xlm-roberta-base}} Both models were trained for the task of masked language modelling for 100 languages.

\begin{table}[t]
\small
	\centering
	\resizebox{.5\textwidth}{!}{
	\begin{tabular}{ll|cccc}
		\toprule[1.pt]
		 Method & Mask & \multicolumn{4}{c}{Dataset} \\
		  & & \englishclo & \english & \german & \japanese \\
		\midrule
		\mbert & mask & 0.76 & 0.69 & 0.68 & 0.48\\
		       & no mask & 0.73 & 0.74 & 0.79 &\secbest{0.49}\\
		\xlmrob & mask & 0.75 & 0.68 & 	0.59 & 0.42\\
		       & no mask & \best{0.79} & \best{0.76} & \best{0.80} & 0.38\\
		\xlm & mask & 0.71 & 0.64 & 0.67 & 0.47 \\
		  & no mask & 0.76 & 0.70 & 0.79 & 0.47 \\
		\midrule
		\textbf{SVM (BoN)} & mask & 0.50&  0.44& 0.47 & \best{0.58}\\
		       & no mask &0.74& 0.70& 0.76& \best{0.58}\\
		\textbf{DT (BoN)} & mask & 0.36 & 0.28 & 0.37& 0.43\\ 
		       & no mask & 0.64 & 0.58& 0.70& 0.48\\
		\textbf{RF (BoN)} & mask & 0.16 & 0.11& 0.20& 0.14\\
		       & no mask & 0.40& 0.34& 0.60 & 0.11\\
		\midrule
		\textbf{SVM (WE)}& mask &0.42& 0.32& 0.40& 0.49 \\
		       & no mask &0.56& 0.49& 0.67& 0.49 \\
		\textbf{DT (WE)}& mask & 0.23& 0.25 & 0.28& 0.42 \\
		       & no mask &0.37 & 0.37& 0.56 & 0.40\\
		\textbf{RF (WE)}& mask & 0.20& 0.08 & 0.17& 0.16\\
		       & no mask &0.26& 0.14& 0.39& 0.14\\
		\bottomrule[1.pt]
	\end{tabular}%
	}
	\caption{MCC for the different CFD Models.}
	\label{tab:cf}
\end{table}

\paragraph{Results}
Here we extend our experiment with clue word masking.
For the transformer-based models we mask the clue words similar to mBERT. 
For the traditional ML methods we remove the clue words from the sentences before tokenization. 

The results with and without masking are reported in \autoref{tab:cf} (F1 and Accuracy are reported in the Supplementary). 
First, we note that masking decreases the performance of all classifiers on all datasets. 
Transformer-based classifiers are the least affected by masking: they are able to learn semantic dependencies from the remaining text. We could also say that transformers are the least affected by the data-selection bias as they do not rely on the clue words.
Traditional ML methods with BoN features are affected by masking the most: they seem to use clue words for discrimination. Interestingly, for these methods the performance drops equally for clue-based \englishclo and enriched \english datasets. This could indicate that in both cases the classifier relies on the clue words.

Overall transformer-based models (especially XLM-RoBERTa) perform the best across all datsets except for \japanese. For \japanese the best performance is obtained by an SVM model with BoN features. 
This could indicate that for Japanese, a language-specific tokenisation works for the lexicalised (BoN) models better than the language-independent subtokenisation methods such as Byte Pair Encoding~\cite[BPE;][]{Sennrich:2016} that are used when training contextualised transformer-based sentence encoders. 
The former preserves more information than the latter at the expense of a sparser and larger feature space~\cite{bollegala2020languageindependent}. 
Transformer-based masked language models on the other hand require subtokenisation as they must use a smaller vocabulary to make the token prediction task efficient~\cite{Yang:2018ab,Li:2019a}. 

In general, unlike the simpler word embedding and bag of words approaches, large pretrained contextualized embeddings maintain high test performance according to the reported evaluation metrics. 
We note that these also converged after a few epochs using a relatively small number of labelled instances, based on the model with the best 5-fold validation accuracy. Hence, contextualized embeddings can identify various context-dependent counterfactuals from a diverse range of reviews using a small number of mini-batch gradient updates of a single linear layer.
Among the different sentence embedding methods compared, the best performance is reported by XLM-RoBERTa.

Between the two baselines, we see that using word embeddings to represent the sentences does not offer clear benefits for traditional ML methods and BoN features are sufficient. However, embedding based methods suffer generally a smaller performance drop when clues are masked. This suggests that embeddings provide a more general and robust representation of counterfactuals in the semantic space than BoN features.

\section{Conclusion}
We annotated a multilingual counterfactual dataset using Amazon product reviews for English, German and Japanese languages.
Experimental results show that our English dataset is compatible with the previously proposed SemEval-2020 Task 5 dataset.
Moreover, the CFD models trained using our dataset are relatively robust against selection bias due to clue phrases.
Simply applying MT on test data results in poor cross-lingual classification performance, indicating the need for language-specific CFD datasets.

\section{Ethical Considerations}
In this work, we annotated a multilingual dataset covering counterfactual statements. 
Moreover, we train CFD models using different sentence representation methods and binary classification algorithms.
In this section, we discuss the ethical considerations related to these contributions.

With regard to the dataset being released, all sentences that are included in the dataset were selected from a publicly available Amazon product review dataset. 
In particular, we do \emph{not} collect or release any additional product reviews as part of this paper.
Moreover, we have manually verified that the sentences in our dataset do not contain any customer sensitive information.
However, product reviews do often contain subjective opinions, which can sometimes be socially biased.
We do not filter out any such biases.

We use two pretrained sentence encoders, mBERT and XLM-RoBERTa,  when training the CFD models. 
It has been reported that pretrained masked language model encode unfair social biases such as gender, racial and religious biases~\cite{bommasani-etal-2020-interpreting}.
Although we have  evaluated ourselves the mBERT and XLM-RoBERTa based CFD models that we use in our experiments, we suspect any social biases encoded in these pretrained masked language models could propagate into the CFD models that we train.
In particular, these social biases could be further amplified during the CFD model training process, if the counterfactual statements in the training data also contain such biases.
Debiasing masked language models is an active research field~\cite{Kaneko:EACL1:2021} and we plan to evaluate the social biases in CFD models in our future work.

\bibliography{CFD}
\bibliographystyle{acl_natbib}


\section*{Supplementary Materials}

\appendix
\section{Fine-tuned multilingual BERT for counterfactual classification}
Given that we select mBERT~\cite{devlin-etal-2019-bert} as the main classification method in the paper, we describe how the original BERT architecture is adapted for fine-tuned for CF classification.

Consider a dataset $D = \{(X_i, y_i)\}_{i=1}^{m}$ for $D \in \mathcal{D}$ and a sample $s:= (X, y)$ where the sentence $X:= (x_1, \ldots x_n)$ with $n$ being the number of words $x \in X$. 
We can represent a word as an input embedding $\vec{x}_w \in \mathbb{R}^d$, which has a corresponding target vector $\vec{y}$. 
In the pre-trained transformer models we use, $X_i$ is represented by 3 types of embeddings; word embeddings ($\mat{X}_w \in \mathbb{R}^{n \times d}$), segment embeddings ($\mat{X}_s \in \mathbb{R}^{n \times d}$) and position embeddings ($\mat{X}_p \in \mathbb{R}^{n \times d}$), where $d$ is the dimensionality of each embedding matrix. 
The self-attention block in a transformer mainly consists of three sets of parameters: the query parameters $\mat{Q} \in \mathbb{R}^{d \times l}$, the key parameters $\mat{K} \in \mathbb{R}^{d \times l}$ and the value parameters $\mat{V} \in \mathbb{R}^{d \times o}$. For 12 attention heads (as in BERT-base), we express the forward pass as follows:

\begin{gather}\label{eq:sa_block}
    \overrightarrow{\mat{X}} = \mat{X}_w + \mat{X}_s + \mat{X}_p  \\
    \overrightarrow{\mat{Z}} := \bigoplus_{i=1}^{12}   \text{softmax}\big(\overrightarrow{\mat{X}} \mat{Q}_{(i)} \mat{K}^{T}_{{(i)}} \overrightarrow{\mat{X}}^{T}\big)\overrightarrow{\mat{X}}\mat{V}_{(i)}  \\
    \overrightarrow{\mathbb{Z}} = \text{Feedforward}(\text{LayerNorm}(\overrightarrow{\mat{Z}} + \overrightarrow{\mat{X}})) \\
        \overleftarrow{\mathbb{Z}} = \text{Feedforward}(\text{LayerNorm}(\overleftarrow{\mat{Z}} + \overleftarrow{\mat{X}})) 
\end{gather}

The last hidden representations of both directions are then concatenated $\mathbb{Z}' :=  \overleftarrow{\mathbb{Z}}  \bigoplus \overrightarrow{\mathbb{Z}'}$ and projected using a final linear layer $\mat{W} \in \mathbb{R}^{d}$ followed by a sigmoid function $\sigma(\cdot)$ to produce a probability estimate $\hat{y}$, as shown in \eqref{eq:cf_layer}. As in the original BERT paper, WordPiece embeddings ~\cite{wu2016google} are used with a vocabulary size of 30,000. Words from (step-3) that are used for filtering the sentences are masked using a \texttt{[PAD]} token to ensure the model does not simply learn to correctly classify some samples based on the association of these tokens with counterfacts. 
A linear layer is then fine-tuned on top of the hidden state, $\vec{h}_{X, \texttt{[CLS]}}$ emitted corresponding to the \texttt{[CLS]} token. This fine-tunable linear layer is then used to predict whether the sentence is counterfactual or not, as shown in \autoref{eq:cf_layer}, where $\cB \subset D$ is a mini-batch and $\mathcal{L}_{ce}$ is the cross-entropy loss.

\begin{align}\label{eq:cf_layer}
 \mathcal{L}_{ce} := \frac{1}{|\cB|} \sum_{(X,y) \in \cB} \vec{y} \log \big( \sigma (\vec{h}_{X,\texttt{[CLS]}} \cdot \mat{W}) \big)
\end{align}

\textbf{Configurations} For the mBERT counterfactual model we use BERT-base, which uses 12 Transformer blocks, 12 self-attention heads with a hidden size of 768. 
The default size of 512 is used for the sentence length and the sentence representation is taken as the final hidden state of the first [CLS] token. 
This model is already pre-trained and we fine-tune a linear layer $\mat{W}$ on top of BERT, which is fed to through a sigmoid function $\sigma$ as $p(c|h) = \sigma(\mat{W}\vec{h})$ where $c$ is the binary class label and we maximize the log-probability of correctly predicting the ground truth label.

\section{Matthews Correlation Coefficient}

Unlike metrics such as F1, MCC accounts for class imbalance and incorporates all correlations within the confusion matrix~\cite{chicco2020advantages}. For MCC, the range is [-1, 1] where 1 represents a perfect prediction, 0 an average random prediction and -1 an inverse prediction.

\begin{equation}\label{eq:mcc}
\textbf{MCC} = \frac{\text{tp} \times \text{tn} - \text{fp} \times \text{fn}}{\sqrt{(\text{tp} + \text{fp})(\text{tp} + \text{fn})(\text{tn} + \text{fp})(\text{tn} + \text{fn})}} 
\end{equation}

\section{Extended version of \autoref{tab:cf}}
We report F1, MCC, and accuracy in \autoref{tab:cf-ext}.

\begin{table*}[t]
\small
	\centering
	\begin{tabular}{ll|ccc|ccc|ccc}
		\toprule[1.pt]
		 Dataset & Mask &  
		 F1 & MCC & Acc &  F1 & MCC & Acc & F1 & MCC & Acc \\
		  && \multicolumn{3}{c}{\textbf{mBERT}} &	\multicolumn{3}{c}{\textbf{XLM-RoBERTa}} & \multicolumn{3}{c}{\textbf{XLM-w/o-Emb}} \\
        \midrule
        \englishclo & mask & 
        \best{0.92} & 0.76 & \best{0.92} &	
        0.91 & 0.75 & 0.90 &
        0.89 & 0.71 & 0.89\\
		 & no mask & 
		 0.89 & 0.73 & 0.89 &	
		 \best{0.92} & \best{0.79} & \best{0.92} &
		 0.91 & 0.76 & 0.90\\
		\midrule
    		 \english & mask & 
		 0.93 & 0.69 & 0.93 &
		 0.93 & 0.68 & 0.92 &
		 0.91 & 0.64 & 0.91\\
		 & no mask & 0.94 & 0.74 & 0.94 &
		 \best{0.95} & \best{0.76} & \best{0.95} & 
		 0.93 & 0.70 & 0.92\\
		\midrule
		\german & mask & 
		 0.86	&	0.68	& 0.86 &
		 0.82	& 	0.59	& 0.82 &
		 0.85	& 	0.67	& 0.85\\
		 & no mask & 
		 0.90 & 0.79 & 0.90 & 	
		 \best{0.91} & \best{0.80} & \best{0.91} & 
		 \best{0.91} & 0.79 & 0.90\\
		 \midrule
		 \japanese & mask & 
		 0.86 & 0.48 & 0.84 &
		 0.81 & 0.42 & 0.78 &
		 0.84 & 0.47 & 0.81\\
		 & no mask &
		 0.85 & \secbest{0.49} & 0.82 &
		 0.86 & 0.38 & 0.85 &
		 0.83 & 0.47 & 0.79\\
		 \midrule
		 \midrule
		 &&  \multicolumn{3}{c}{\textbf{SVM (BoN)}} &\multicolumn{3}{c}{\textbf{DT (BoN)}} & \multicolumn{3}{c}{\textbf{RF (BoN)}}\\
		 \englishclo & mask & 
		 0.83 & 0.50 & 0.83 &
		 0.80 & 0.36 & 0.82 &
		 0.73 & 0.16 & 0.81\\
		 & no mask &
		 0.91 & 0.74 & 0.91 &
		 0.88 & 0.64 & 0.89 &
		 0.80 & 0.40 & 0.84\\
		 \midrule
		 \english & mask & 0.89 & 0.44 & 0.88 &
		 0.87 & 0.28 & 0.89 &
		 0.85 & 0.11 & 0.89\\
		 & no mask & 0.94 & 0.70 & 0.94 &
		 0.92 & 0.58 & 0.93 &
		 0.87 & 0.34 & 0.90\\
		 \midrule
		 \german & mask & 
		 0.76 & 0.47 & 0.75 &
		 0.73 & 0.37 & 0.75 &	
		 0.66 & 0.20 & 0.71\\
		 & no mask & 
		 0.89 & 0.76 & 0.89 &
		 0.87 & 0.70 & 0.87 &	
		 0.82 & 0.60 & 0.83
		\\
		\midrule
		\japanese & mask &
		\best{0.91} & \best{0.58} & 0.91 &
		0.90 & 0.43 & 0.91 &
		0.85 & 0.14 & 0.89\\
		& no mask & 
        \best{0.91} & \best{0.58} & 0.91 &
	    0.90 & 0.48 & \best{0.92} &
	    0.85 & 0.11 & 0.89
		\\\midrule
		&&  \multicolumn{3}{c}{\textbf{SVM (WE)}} &\multicolumn{3}{c}{\textbf{DT (WE)}} & \multicolumn{3}{c}{\textbf{RF (WE)}}\\
		\englishclo & mask & 
		0.78 & 0.42 & 0.77 &
		0.77 & 0.23 & 0.79 &
		0.74 & 0.20 & 0.81\\
		& no mask &
		0.84 & 0.56 & 0.82 &
		0.81 & 0.37 & 0.82 &
		0.76 & 0.26 & 0.82 \\
		\midrule
		\english & mask & 
		 0.80 & 0.32 & 0.76&
		 0.87 & 0.25 & 0.90&	
		 0.84 & 0.08 & 0.89\\
		 & no mask &
		 0.86 & 0.49 & 0.84&
		 0.89 & 0.37 & 0.90&
		 0.85 & 0.14 & 0.89\\
		 \midrule
		 \german & mask & 
		 0.71 & 0.40 & 0.70&
		 0.70 & 0.28 & 0.72&
		 0.65 & 0.17 & 0.70\\
		 & no mask & 
		 0.84 & 0.67 & 0.84&
		 0.81 & 0.56 & 0.82&
		 0.73 & 0.39 & 0.76
		\\
		\midrule
		\japanese & mask &
		0.87 & 0.49 & 0.84&
		0.90 & 0.42 & 0.91&
		0.85 & 0.16 & 0.90\\
		& no mask & 
		0.86 & 0.49 & 0.84&
		0.89 & 0.40 & 0.91 &
		0.85 & 0.14 & 0.89\\
		\bottomrule[1.pt]
	\end{tabular}%
	\caption{F1, Matthew's Correlation Coefficient \& Accuracy for the different CFD Models.}
	\label{tab:cf-ext}
	\vspace{-5mm}
\end{table*}

\section{Examples of Incorrect Predictions}

\begin{table*}[ht]
\centering

\resizebox{1.\linewidth}{!}{
\begin{tabular}{|l |c}

\toprule
\hline
\multicolumn{1}{c}{Misclassifications of Reviews Containing No Counterfactuals} & Models\\
\hline
If you workout regularly, an extra set of 'expendable' earbuds like these is a must-have. & B XR X\\ 
I put over 500 songs on it the first day and still have around 17 GB left, probably could have done with a much smaller one.& B XR X\\ 
If you have a similar build compared to mine, buy this shirt without hesitation.& B XR X\\ 
If they ever need replacing I would definitely buy these again. & B XR X\\ 
If this device for whatever reason fails within a year or two, I think I would look to buy the same machine again. & B XR X\\ 
I was hoping she would be able to grow in it but it fits her now with no room to grow.& B XR X\\ 
I must have read reviews on about 20 different models. & B XR X\\ 
Because it is fleece, if you are in the US, I would suggest a second cool water rinse with a touch of fabric softener. & B XR X\\ 
There are ways to get it like you want it but its not as easy as it could have been. & B XR X\\ 
Could be improve with a size adjustment and chin strap. & B XR X\\ 
If you need more desk space and have a location where you can use a wall mount for your monitors, this thing is the way to go. & B XR X\\ 
It should be about \$20 cheaper to make it worth while.  & B XR X\\ 
\midrule
\multicolumn{2}{c}{Misclassifications of Counterfactual Reviews} \\
\hline
\midrule
At the end of a series like The Wheel of Time, it might be appropriate to lament the loss of familiar characters.& B XR X\\ 
You would have to be 5'10 and super thin to fit into these. & B XR X\\ 
From the picture the dress looks like it should be long enough for someone at lease 5' 6. & B XR X\\ 
To say "the usual awesome Stephen King novel" would be an understatement. & B XR X\\ 
I don't like to go into the plot a lot unless the blurb doesn't represent the book fairly. & B XR X\\
I've thought about it, and I guess that's because what happened to the characters in Missing are stuff that I could imagine happening to me as well. & B XR \\
For the price that this particular seller charged for this T-shirt, the material SHOULD be HEAVY-DUTY. & B X\\
If one can put aside their religious beliefs about heaven and hell I think they will find this to be something they've always known deep inside about the afterlife. & XR X\\
If you think leakage is a problem it really isn't they are as bad as a pair of ear-buds.  & XR X\\
\bottomrule

\end{tabular}}
    \caption{Qualitative Examples of Incorrect Predictions from Fine-tuned BERT}
    \label{tab:qual_examps}
\end{table*}

~\autoref{tab:qual_examps} shows examples of misclassifications given by transformer models. The second column indicates which of the remaining transformer models misclassified each review where B=mBERT, XR=XLM-RoBERTa, X=XLM without embedding.

\section{Hardware Used}
All transformer, RNN and CNN models were trained using a GeForce NVIDIA GTX 1070 GPU which has 8GB GDDR5 Memory.

\section{Model Configuration and Hyperparameter Settings}
BERT-base uses 12 Transformer blocks, 12 self-attention heads with a hidden size of 768.  The default size of 512 is used for the sentence length and the sentence representation is taken as the final hidden state of the first [CLS] token. A fine-tuned linear layer $\mat{W}$ is used on top of BERT-base, which is fed to through a sigmoid function $\sigma$ as $p(c|h) = \sigma(\mat{W}\vec{h})$ where $c$ is used to calibrate the class probability estimate and we maximize the log-probability of correctly predicting the ground truth label.

\autoref{tab:pretrain_hyperparam_transformer} shows the pretrained model configurations that were already predefined before our experiments. The number of (Num.) hidden groups  here are the number of groups for the hidden layers where parameters in the same group are shared. The intermediate size is the dimensionality of the feed-forward layers of the the Transformer encoder.
The `Max Position Embeddings' is the maximum sequence length that the model can deal with.

\begin{table}[ht]
	\centering
	\resizebox{.5\textwidth}{!}{%
	\begin{tabular}{l|c|c|c}
		\toprule[1.pt]
		 Hyperparameters & \mbert & \xlmrob & \xlm \\
		\hline
		Vocab Size & 119547 & 250002 & 119547\\
        Max Pos. Embeddings & 512 & 514 & 514\\
		Hidden Size & 3072 & 3072 & 3072 \\
		Encoder Size & 768 & 768 & 768 \\
        Num. Hidden Layers & 12 & 12 & 12\\
        Num. Hidden Groups & 1 & 1 & 1 \\
        Num. Attention Heads & 12 & 12 & 12 \\

        Hidden Activations & GeLU & GeLU & GeLU \\
        Layer Norm. Epsilon & $ 10^{-12}$  & $ 10^{-12}$ & $ 10^{-12}$\\
        Fully-Connected Dropout Prob. & 0.1  & 0.1 & 0.1\\

        Attention Dropout Prob. & 0  & 0 & 0\\
	\bottomrule[1.pt]
	\end{tabular}%
	}
	\caption{Final Transformer Model Hyperparameter Settings}
	\label{tab:pretrain_hyperparam_transformer}
\end{table}

We now detail the hyperparameter settings for transformer models and the baselines. We note that all hyperparameter settings were performed using a manual search over development data. 

\subsection{Transformer Model Hyperparameters}
We did not change the original hyperparameter settings that were used for the original pretraining of each transformer model. The hyperparameter settings for these pretrained models can be found in the class arguments python documentation in each configuration python file in the \url{https://github.com/huggingface/transformers/blob/master/src/transformers/} e.g configuration\_.py and are also summarized in \autoref{tab:pretrain_hyperparam_transformer}.

\begin{table}[ht]
	\centering
	\resizebox{.48\textwidth}{!}{%
	\begin{tabular}{l|c|c|c}
		\toprule[1.pt]
		 Hyperparameters & \mbert & \xlmrob & \xlm\\
		\hline
		Seed & 1234 & 1234 & 1234\\
        Learning rates &  $10^{-5}$ &  $10^{-5}$ & $50^{-5}$\\
		Max Seq. Length & 256 & 256 & 256\\
        Max Train Epochs & 50 & 50 & 50\\
        Warmup Proportion & 0.1  & 0.1 & 0.1\\
        Classifier Dropout Prob. & 0.2 & 0.2 & 0.2 \\
        Adam eps & $10^{-8}$ & $10^{-8}$ & $10^{-8}$\\
	\bottomrule[1.pt]
	\end{tabular}%
	}
	\caption{Final Transformer Model Hyperparameter Settings}
	\label{tab:tune_hyperparam_transformer}
\end{table}

For fine-tuning transformer models, we manually tested different combinations of a subset of hyperparameters including the learning rates $\{50^{-4}, 10^{-5}, 50^{-5}\}$, batch sizes $\{16, 32, 128\}$, warmup proportion $\{ 0, 0.1\}$ and $\epsilon$ which is a hyperparameter in the adaptive momentum (adam) optimizer. 
Please refer to the huggingface documentation at \url{https://github.com/huggingface/transformers} for further details on each specific model e.g at \url{https://github.com/huggingface/transformers/blob/master/src/transformers/modeling_bert.py}, and also for the details of the architecture for BertForSequenceClassification pytorch class that is used for our sentence classification and likewise for the remaining models.

Fine-tuning all language models with a sentence classifier took less than two and half hours for all models. 
For example, for the largest transformer model we used, BERT, the estimated average runtime for a full epoch with batch size 16 (of 2, 682 training samples) is 184.13 seconds. In the worst case, if the model does not already converge early and all 50 training epochs are carried out, training lasts for 2 hour and 30 minutes. 

\subsection{Baseline Hyperparameters}

\paragraph{SVM Classifier:} 
A radial basis function was used as the nonlinear kernel, tested with an $\ell_2$ regularization term settings of  $C=\{0.01, 0.1, 1 \}$, while the kernel coefficient $\gamma$ is autotuned by the scikit-learn python package and class weights are used inversely proportional to the number of samples in each class. To calibrate probability estimates for AUC scores, we use Platt's scaling~\citep{platt1999probabilistic}.

\paragraph{Decision Tree and Random Forest Classifiers:}
We use 20 decision tree classifiers with no restriction on tree depth and the minimum number of samples required to split an internal node is set to 2. The criterion for splitting nodes is the Gini importance~\citep{gini1912variabilita}.

\newpage
\section{Further Details on the Datasets}
\paragraph{Review categories represented in the datasets and clue words breakdown:}
Below we list the breakdown of product categories for each dataset in the format
``category (total number of review sentences from the category / number of counterfactual examples/ number of non-counterfactual examples)''.

Dataset \english contains review sentences from 4 product categories:
Apparel (2500 / 297 / 2203),
Digital\_Ebook\_Purchase (2500 / 287 / 2213),
Electronics (2500 / 213 / 2287),
Home (2500 / 233 / 2267).

Dataset \german contains review sentences from 20 categories:
Automotive (47 / 31 / 16),
Baby (99 / 80 / 19),
Camera (816 / 597 / 219),
Digital\_Ebook\_Purchase (426 / 259 / 167),
Digital\_Video\_Download (1297 / 961 / 336),
Electronics (7 / 5 / 2),
Home Entertainment (94 / 62 / 32),
Home Improvement (87 / 54 / 33),
Kitchen (20 / 10 / 10),
Lawn and Garden (47 / 34 / 13),
Luggage (21 / 9 / 12),
Music (1297 / 909 / 388),
Musical Instruments (162 / 113 / 49),
Office Products (40 / 25 / 15),
PC (1297 / 873 / 424),
Personal\_Care\_Appliances (56 / 36 / 20),
Sports (5 / 3 / 2),
Toys (378 / 216 / 162),
Watches (186 / 126 / 60),
Wireless (618 / 437 / 181).

Dataset \japanese contains review sentences from 18 categories:
Automotive (191 / 19 / 172),
Baby (182 / 6 / 176),
Camera (490 / 67 / 423),
Digital\_Ebook\_Purchase (490 / 22 / 468),
Digital\_Video\_Download (490 / 49 / 441),
Electronics (490 / 43 / 447),
Home (102 / 16 / 86),
Home Entertainment (227 / 34 / 193),
Home Improvement (221 / 29 / 192),
Kitchen (221 / 23 / 198),
Music (490 / 21 / 469),
Musical Instruments (490 / 42 / 448),
PC (490 / 61 / 429),
Shoes (490 / 52 / 438),
Sports (466 / 39 / 427),
Toys (490 / 53 / 437),
Watches (490 / 37 / 453),
Wireless (490 / 54 / 436).

The clue phrases for English, German and Japanese are shown respectively in \autoref{tab:en:clue_words}, \autoref{tab:de:clue_words} and \autoref{tab:jp:clue_words}.

\begin{table}[h!]
	\setlength{\tabcolsep}{0pt}
	\centering
	\resizebox{.48\textwidth}{!}{%
	\begin{tabular*}{0.5\textwidth}{@{\extracolsep{\fill}} lccccc}
		\toprule 
		Clue phrase & $N_P$ & $N_N$ & $f_{P}$&$f_{N}$ & $f_{data}$  \\
		\midrule
		without&25&571&2.42&6.36&5.96\\
		doesn't&11&569&1.06&6.34&5.8\\
		wanted&18&512&1.74&5.70&5.3\\
		would be&100&389&9.70&4.33&4.89\\
		would have&281&114&27.2&1.27&3.95\\
		wish&305&81&29.6&0.90&3.86\\
		couldn't&11&289&1.06&3.22&3.0\\
		won't&9&274&0.87&3.05&2.83\\
		must&13&258&1.26&2.87&2.71\\
		haven't&5&208&0.48&2.31&2.13\\
		instead of&18&132&1.74&1.47&1.5\\
		should be&20&126&1.94&1.40&1.46\\
		came with&9&128&0.87&1.42&1.37\\
		could have&100&36&9.70&0.40&1.36\\
		should have&106&19&10.2&0.21&1.25\\
		miss&6&116&0.58&1.29&1.22\\
		could be&19&103&1.84&1.14&1.22\\
		except&7&115&0.67&1.28&1.22\\
		comes with&1&80&0.09&0.89&0.81\\
		none&2&68&0.19&0.75&0.7\\
		missing&3&56&0.29&0.62&0.59\\
		if it was&21&30&2.03&0.33&0.51\\
		might have&10&20&0.97&0.22&0.3\\
		wished&18&4&1.74&0.04&0.22\\
		had not&3&13&0.29&0.14&0.16\\
		if it were&13&3&1.26&0.03&0.16\\
		if it had&10&3&0.97&0.03&0.13\\
		wishing&5&4&0.48&0.04&0.09\\
		had thought&3&4&0.29&0.04&0.07\\
		\midrule
		Total & 954&4069&92.6&45.3&50.2\\
		\bottomrule
	\end{tabular*}
	}
		\caption{English clue words (statistics for \english dataset). $N_P$ (and $N_N$) is the number of positive (and negative) examples with the clue word. $f_P$ (and $f_N$) is the percent of positive (negative) examples with the clue word. $f_{data}$ is the frequency of the clue word in the dataset.}
		\label{tab:en:clue_words}
\end{table}

\begin{table}[th!]
	\setlength{\tabcolsep}{0pt}
	\centering
	\resizebox{.48\textwidth}{!}{%
	\begin{tabular*}{0.5\textwidth}{@{\extracolsep{\fill}} lccccc}
		\toprule 
		Clue phrase & $N_P$ & $N_N$ & $f_{P}$&$f_{N}$ & $f_{data}$  \\
		\midrule
		hätte&1804&11&37.2&0.50&25.9\\
		wäre&1397&22&28.8&1.01&20.2\\
		könnte&1143&28&23.6&1.29&16.7\\
		müssen&122&479&2.52&22.1&8.58\\
		fehlt&111&429&2.29&19.8&7.71\\
		wenn es&296&227&6.11&10.5&7.47\\
		statt&107&200&2.21&9.25&4.38\\
		außer&52&184&1.07&8.51&3.37\\
		wünschen&115&80&2.37&3.70&2.78\\
		müsste&174&13&3.59&0.60&2.67\\
		wird nicht&15&167&0.30&7.73&2.6\\
		eigentlich nicht&51&119&1.05&5.50&2.42\\
		dürfen&34&63&0.70&2.91&1.38\\
		vermisse&10&55&0.20&2.54&0.92\\
		gewollt&4&33&0.08&1.52&0.52\\
		wünschte&25&11&0.51&0.50&0.51\\
		verpassen&13&22&0.26&1.01&0.5\\
		konnte nicht&4&27&0.08&1.25&0.44\\
		hatte nicht&6&11&0.12&0.50&0.24\\
		haben nicht&1&14&0.02&0.64&0.21\\
		könnte sein&6&0&0.12&0.0&0.08\\
		hatte gedacht&2&4&0.04&0.18&0.08\\
		nicht hätte&5&0&0.10&0.0&0.07\\
		anstelle von&1&2&0.02&0.09&0.04\\
		hätte haben können&0&0&0.0&0.0&0.0\\
		sollte haben&0&0&0.0&0.0&0.0\\
		wenn es hatte&0&0&0.0&0.0&0.0\\
		\midrule
		Total & 4840&2160&100.&100.&100.\\
		\bottomrule
	\end{tabular*}
	}
	\caption{German clue words. $N_P$ (and $N_N$) is the number of positive (and negative) examples with the clue word. $f_P$ (and $f_N$) is the percent of the clue word in the dataset.}
	\label{tab:de:clue_words}
\end{table}

\begin{CJK}{UTF8}{min}
	\begin{table}[th!]
		\setlength{\tabcolsep}{0pt}
		\centering
		\resizebox{.48\textwidth}{!}{%
		\begin{tabular*}{0.5\textwidth}{@{\extracolsep{\fill}} lccccc}
			\toprule 
			Clue phrase & $N_P$ & $N_N$ & $f_{P}$&$f_{N}$ & $f_{data}$  \\
			\midrule
			思います&84&1556&12.5&24.5&23.4\\
			れば&258&805&38.6&12.7&15.1\\
			なら&84&841&12.5&13.2&13.2\\
			でしょう&28&633&4.19&9.99&9.44\\
			良かった&63&324&9.44&5.11&5.52\\
			思う&32&344&4.79&5.43&5.37\\
			いいです&17&282&2.54&4.45&4.27\\
			よかった&53&239&7.94&3.77&4.17\\
			思った&12&251&1.79&3.96&3.75\\
			良いです&9&233&1.34&3.67&3.45\\
			思いました&22&204&3.29&3.22&3.22\\
			だろう&12&211&1.79&3.33&3.18\\
			もっと&94&116&14.0&1.83&3.0\\
			もう少し&131&64&19.6&1.01&2.78\\
			良いと&21&154&3.14&2.43&2.5\\
			ほうが&18&155&2.69&2.44&2.47\\
			いいと&14&143&2.09&2.25&2.24\\
			助か&7&119&1.04&1.87&1.8\\
			べき&13&102&1.94&1.61&1.64\\
			さらに&6&108&0.89&1.70&1.62\\
			欲しかった&42&44&6.29&0.69&1.22\\
			としても&2&67&0.29&1.05&0.98\\
			いいかも&6&58&0.89&0.91&0.91\\
			ならば&8&52&1.19&0.82&0.85\\
			更に&7&51&1.04&0.80&0.82\\
			たかった&10&46&1.49&0.72&0.8\\
			思っていました&5&48&0.74&0.75&0.75\\
			できれば&24&28&3.59&0.44&0.74\\
			良いのですが&6&35&0.89&0.55&0.58\\
			だったら&15&26&2.24&0.41&0.58\\
			良いかも&2&39&0.29&0.61&0.58\\
			おもいます&3&35&0.44&0.55&0.54\\
			ほしかった&25&11&3.74&0.17&0.51\\
			よいと&5&29&0.74&0.45&0.48\\
			いいのですが&6&26&0.89&0.41&0.45\\
			いいかな&7&22&1.04&0.34&0.41\\
			思ってた&1&25&0.14&0.39&0.37\\
			よいです&0&25&0.0&0.39&0.35\\
			いいな&7&18&1.04&0.28&0.35\\
			たらな&3&20&0.44&0.31&0.32\\
			いいのに&8&8&1.19&0.12&0.22\\
			\midrule
			\multicolumn{6}{r}{Continued on next column}\\
			\bottomrule
		\end{tabular*}
		}
			\caption{Japanese clue words. $N_P$ (and $N_N$) is the number of positive (and negative) examples with the clue word. $f_P$ (and $f_N$) is the fraction of positive (negative) examples with the clue word. $f_{data}$ is the frequency of the clue word in the dataset.}
			\label{tab:jp:clue_words}
	\end{table}
\end{CJK}

\begin{CJK}{UTF8}{min}
	\begin{table}[th!]
		\setlength{\tabcolsep}{0pt}
		\centering
		\resizebox{.48\textwidth}{!}{%
		\begin{tabular*}{0.5\textwidth}{@{\extracolsep{\fill}} lccccc}
		    \toprule 
		    \multicolumn{6}{r}{Continued from previous page}\\
			\midrule 
			Clue phrase & $N_P$ & $N_N$ & $f_{P}$&$f_{N}$ & $f_{data}$  \\
			\midrule
			良いのでは&3&13&0.44&0.20&0.22\\
			良いかな&2&10&0.29&0.15&0.17\\
			いいのでは&1&10&0.14&0.15&0.15\\
			したかった&0&10&0.0&0.15&0.14\\
			おもう&1&7&0.14&0.11&0.11\\
			いいんですが&0&7&0.0&0.11&0.1\\
			良いのだが&4&3&0.59&0.04&0.1\\
			だったのに&1&6&0.14&0.09&0.1\\
			おもった&0&6&0.0&0.09&0.08\\
		    おもいました&1&4&0.14&0.06&0.07\\
			良いな&2&3&0.29&0.04&0.07\\
			良かったのに&3&2&0.44&0.03&0.07\\
			よいかな&2&2&0.29&0.03&0.05\\
			よいのでは&0&4&0.0&0.06&0.05\\
			よいのですが&2&2&0.29&0.03&0.05\\
			たすかり&1&2&0.14&0.03&0.04\\
			よいかも&0&3&0.0&0.04&0.04\\
			よかったのに&2&0&0.29&0.0&0.02\\
			よいのに&2&0&0.29&0.0&0.02\\
			ところだった&0&1&0.0&0.01&0.01\\
			よいな&1&0&0.14&0.0&0.01\\
			良いのに&0&1&0.0&0.01&0.01\\
			おもっていました&0&0&0.0&0.0&0.0\\
			おもってた&0&0&0.0&0.0&0.0\\
			たすかった&0&0&0.0&0.0&0.0\\
			たすかる&0&0&0.0&0.0&0.0\\
			よいんですが&0&0&0.0&0.0&0.0\\
			あれば&0&0&0.0&0.0&0.0\\
			\midrule
			Total & 667&6333&100.&100.&100.\\
			\bottomrule
		\end{tabular*}
		}
	\end{table}
\end{CJK}

\end{document}